# Improving Simulation Regression Efficiency using a Machine Learning-based Method in Design Verification

Deepak Narayan Gadde, Infineon Technologies Dresden GmbH & Co. KG, Dresden, Germany
(*Deepak.Gadde@infineon.com*)

Sebastian Simon, Infineon Technologies Dresden GmbH & Co. KG, Dresden, Germany
(*Sebastian.Simon@infineon.com*)

Djones Lettnin, Infineon Technologies AG, Munich, Germany
(*Djones.Lettnin@infineon.com*)

Thomas Ziller, Cadence Design Systems GmbH, Munich, Germany
(*thomasz@cadence.com*)

*Abstract*— The verification throughput is becoming a major challenge bottleneck, since the complexity and size of SoC designs are still ever increasing. Simply adding more CPU cores and running more tests in parallel will not scale anymore. This paper discusses various methods of improving verification throughput: ranking and the new machine learning (ML) based technology introduced by Cadence i.e. Xcelium ML. Both methods aim at getting comparable coverage in less CPU time by applying more efficient stimulus. Ranking selects specific seeds that simply turned out to come up with the largest coverage in previous simulations, while Xcelium ML generates optimized patterns as a result of finding correlations between randomization points and achieved coverage of previous regressions. Quantified results as well as pros & cons of each approach are discussed in this paper at the example of three actual industry projects. Both Xcelium ML and Ranking methods gave comparable compression & speedup factors around 3 consistently. But the optimized ML based regressions simulated new random scenarios occasionally producing a coverage regain of more than 100%. Finally, a methodology is proposed to use Xcelium ML efficiently throughout the product development.

## I. INTRODUCTION

At present, due to the growth in various usage requirements and demands from customers to the semiconductor manufacturers, the complexity in hardware design is rapidly rising. This elevates the efforts for design verification engineers to verify such complex designs. Generally, most of the verification flows include a large number of test simulations with random stimuli to prove the design in all significant scenarios. Such scenarios are simulated by constrained random tests and can be tracked by coverage metrics in the testbench. For an ideal verification sign-off, 100% coverage is required. A lot of efforts and time are required to run such huge simulation regressions and to analyze them on a regular basis. The major problem is that these regressions consists of redundant simulations which do not improve coverage. This issue no longer can be addressed by simply adding more CPUs and running more tests in parallel. These large size regressions demand high number of simulator resources and require long turn-around times.

A methodology to optimize these simulation regressions and to speed up the design verification process throughout the product development is investigated in this paper using "Xcelium ML" [1]. This tool uses Machine Learning technology to optimize simulation regressions and to produce a well condensed regression. This optimized regression can be used to achieve similar coverage as the original regression and to find design bugs quickly by simulating the corner case scenarios with the existing randomized testbench.

This work has been developed in the project VE-VIDES (project label 16ME0243K) which is partly funded within the Research Programme ICT 2020 by the German Federal Ministry of Education and Research (BMBF)

## II. RELATED WORK

Sundeep Srinivasan et al. presented a paper [2] showing significant results in reducing time to find hard to reach conditions in design verification using machine learning techniques. The approach uses Reinforcement Learning to produce better stimuli to reach the desired coverage proving that ML can do better than random.

David Crutchfield et al. proposed an approach [3] to eliminate tests and to improve coverage during the simulation regressions. It is not based on any artificial intelligence techniques. It rather relies on a well-developed automated algorithm to find an optimized test list with certain random seed values and uses ranking and test weighted flow.

In a recent paper [4], Tim Blackmore et al. showed the results that the novelty detection using an Autoencoder can be used to select tests that are more likely to add to coverage. Their proposed approach is applied to the verification of a Radar Signal Processing Unit (RSPU) design giving noteworthy results in coverage. They have concluded that the coverage of RSPU could be closed or very nearly closed using novelty selection whilst simulating 60% fewer tests than with existing constrained random generation.

Most of the ML methods published require the engineer to put efforts in collecting the data to train the model and to analyze it. A new methodology is proposed in this paper using an EDA tool reducing the efforts for the verification engineers.

## III. INTRODUCTION TO THE METHODOLOGIES AND SUCCESS METRICS

### A. Ranking

A Ranking feature is offered by various verification management tools from different EDA vendors. As testbenches are written using Constrained Random Verification (CRV) methodology, there is a randomness problem. Ranking is used to choose the tests from the simulation regression according to their contribution in terms of coverage improvement, thus providing the most efficient test suite. Input for this method is a simulation regression with random seed values. The result is a compressed regression with certain tests with certain seed values while eliminating the redundant simulation runs that do not add any additional coverage. The flow for the ranking method is shown below in Figure 1.

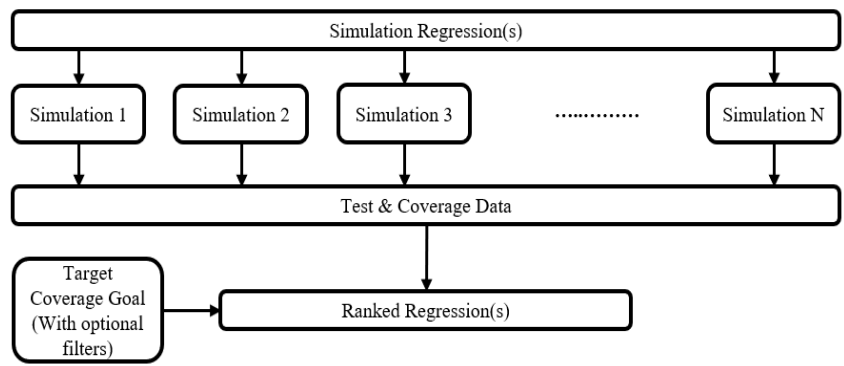

Figure 1. Flow for Ranking & Selection

### B. Xcelium ML

The Xcelium ML tool flow is shown in Figure 2. It takes a set of regressions as input and collects the coverage, as well as control data during each of the simulations, which serve as training data for the machine learning models. The tool uses the unique tests and front-end random variables as the input layer for the neural network, with the output layer being the coverage bin. This output layer is an abstraction of a bin where one output could represent the number of bins. The depth of the complete neural net depends on the bin. So, for each of the simulation regression multiple neural networks might get generated.

After these ML neural networks are trained, the goals are to be set by the user to produce the optimized ML regression (e.g. a goal can be a regression with a dedicated number of tests or one or more target coverage values for different parts or the whole design). It is possible to give different goals as input for these trained ML networks to generate multiple optimized regression sets. These trained ML models predict under which input state conditions the provided design gets into which specific state. Thus, the models will get to know, which tests with which random variables are needed to get the design into some specific state of operation. The optimized regression selects relevant tests and comes with additional constraining to ensure the relevant states of operation are reached.

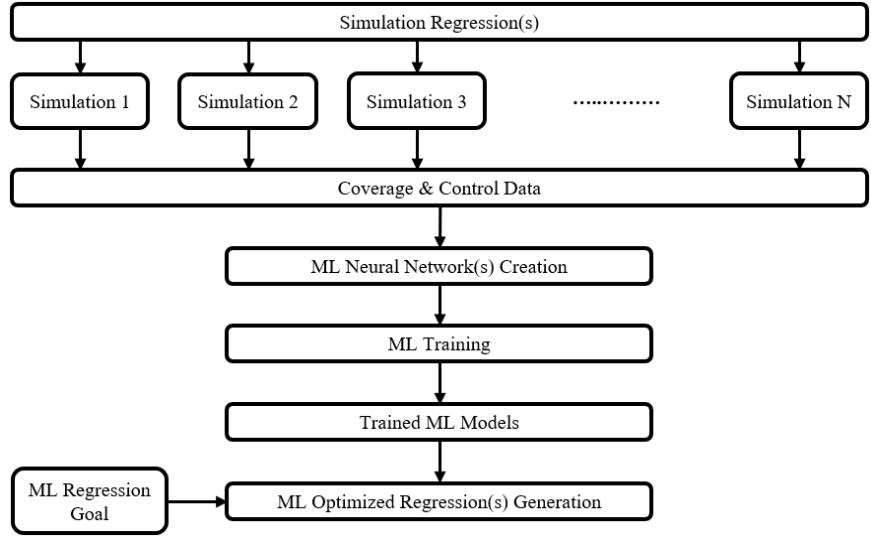

Figure 2. Xcelium ML tool flow

### C. Metrics to Compare the Methodologies

Ranking and Xcelium ML aim at producing comparable coverage of the original regression in less CPU time by applying more efficient stimulus. Both methods are applied on three real industry projects and the results are discussed in Section IV. The key metrics used for comparison are coverage regain, compression in total runs, and optimization in CPU runtime. These metrics are calculated using the formulae given below.

$$Coverage\ regain = \frac{Coverage\ of\ the\ optimized\ regression}{Coverage\ of\ the\ original\ regression} \times 100$$

$$Compression\ Factor\ in\ total\ runs = \frac{Total\ runs\ in\ the\ original\ regression}{Total\ runs\ in\ the\ optimized\ regression}$$

$$Compression\ Factor\ in\ CPU\ runtime = \frac{Total\ CPU\ runtime\ of\ the\ original\ regression}{Total\ CPU\ runtime\ of\ the\ optimized\ regression}$$

Coverage regain gives the stats on reproducing the coverage of the original regression with an optimized regression generated by Xcelium ML or ranking techniques. The time savings in reproducing the original coverage can be determined by the metric "Compression factor in CPU runtime". This metric can also be defined as Speedup factor. Compression Factor in total runs answers the question of how strongly the number of runs in a regression could be reduced.

## IV. RESULTS

Xcelium ML is applied to optimize the existing regressions used by the verification team on multiple projects & designs. The results produced by both methods during each of the project is explained in the following subsections. Later, ranking results are compared with Xcelium ML results to understand the performance of both methodologies.

A. *Microprocessor IP*

The Microprocessor IP is an Automotive Safety Integrity Level (ASIL) D category product designed to fulfilling the ISO26262 standard. This IP is also a Safety Element out of Context (SEooC) [5] based on the Harvard architecture. The RTL design of this IP is implemented in VHDL. The testbench for the respective design is modeled using SystemVerilog-UVM. It has some internal methodology for configuration setup and also employ CRV.

The complete testbench has 26 tests in total to verify the design functionality. These 26 tests run 10 times each with random seeds, resulting in a total of 260 runs in Stage I regression. Later in the design cycle, after further development of both design and testbench, these 26 tests were simulated 20 times each producing a regression of 520 runs in Stage II. The results of the Xcelium ML optimized regression are shown in Table I as follows.

Table I. Results produced by Xcelium ML on Microprocessor IP

|          | Original Runs | Original Runtime CPU hours | Optimized Runs | Optimized Runtime CPU hours | Compression Factor in Runs | Coverage Regain | Compression Factor in Runtime |
|----------|---------------|----------------------------|----------------|-----------------------------|----------------------------|-----------------|-------------------------------|
| Stage I  | 260           | 12                         | 26             | 0.7                         | 10                         | 99.90%          | 17.14                         |
| Stage II | 520           | 48                         | 29             | 5.2                         | 17.93                      | 99.89%          | 9.23                          |

The ranking feature is applied on the same regressions to produced ranked regressions as shown in Table II. The results produced by ranking show higher compression factors in comparison with the results produced by Xcelium ML. Comparing both Xcelium ML and ranking at Stage I, ranking has chosen the best test-seed pairs to regenerate the same coverage 11 times faster than the actual regression, whereas the Xcelium ML could regenerate 99.90% of the original coverage 17 times faster.

At Stage II, specifically, some new bins have been hit by the ML optimized regression in comparison with the original regression. Ranked regressions generally hit exactly the same bins as the original regressions.

Table II. Results produced by Ranking on Microprocessor IP

|          | Original Runs | Original Runtime CPU hours | Ranked Runs | Ranked Regression Runtime CPU hours | Compression Factor in Runs | Coverage Regain | Compression Factor in Runtime |
|----------|---------------|----------------------------|-------------|-------------------------------------|----------------------------|-----------------|-------------------------------|
| Stage I  | 260           | 12                         | 8           | 1.1                                 | 32.5                       | 100%            | 10.91                         |
| Stage II | 520           | 48                         | 9           | 1.4                                 | 57.77                      | 100%            | 34.29                         |

B. *Mixed Signal SoC*

This design is a complex mixed signal SoC. The RTL is designed in VHDL, Verilog, SystemVerilog, and AMS languages while the testbench is written in SystemVerilog using UVM and CRV methodologies verifying both digital and analog-mixed signal blocks of the design. The regression of this SoC consists of 5124 simulations in total (356 distinct tests). These tests are classified into 7 unique configurations depending on instantiation of a CPU and, Verification IP, and type of used flash memory. The configurations used can be seen in the following Table III.

Table III. Various configurations of mixed signal SoC

| Configurations | CPU | Verification IP | Flash   |
|----------------|-----|-----------------|---------|
| P1             | Yes | No              | Simple  |
| P2             | Yes | Yes             | Complex |
| P3 | Yes | Yes | Simple |
| P4 | No | No | Complex |
| P5 | No | No | Simple |
| P6 | No | Yes | Complex |
| P7 | No | Yes | Simple |

The tool is applied to the above configurations producing the results mentioned in Table IV. The total regression suite has 5124 test runs which are optimized to 1605 test runs with Xcelium ML regaining 99.42% of the original coverage. The optimization factor on average is around 3 (both reduction in number of runs and CPU time). The ML based regression is able to re-achieve the original coverage faster. Especially, the configurations P2 and P7 are able to regain more than 100% coverage with less simulations compared to the original regression. This shows that randomized regressions produced with ML method can even hit new coverbins in less time and with fewer runs.

Table IV. Results produced by Xcelium ML on mixed signal SoC

| Configuration | Original Runs | Original Runtime CPU hours | Optimized Runs | Optimized Runtime CPU hours | Compression Factor in Runs | Coverage Regain | Compression Factor in Runtime |
|---|---|---|---|---|---|---|---|
| P1 | 297 | 0.200 | 110 | 0.151 | 2.70 | 99.86% | 1.32 |
| P2 | 415 | 0.562 | 188 | 0.337 | 2.20 | 101.2% | 1.66 |
| P3 | 2959 | 2.813 | 786 | 0.785 | 3.76 | 97.96% | 3.58 |
| P4 | 140 | 0.213 | 116 | 0.133 | 1.20 | 98.07% | 1.60 |
| P5 | 1193 | 1.336 | 318 | 0.356 | 3.75 | 99.48% | 3.75 |
| P6 | 10 | 0.007 | 3 | 0.003 | 3.33 | 98.19% | 2.33 |
| P7 | 110 | 0.074 | 84 | 0.070 | 1.30 | 101.3% | 1.05 |
| Total | 5124 | 5.205 | 1605 | 1.835 | 3.19 | 99.42% | 2.83 |

Ranking technique is also applied on the same regressions producing the results in Table V. Here the overall compression factors are actually higher than the compression produced by Xcelium ML. But, as discussed earlier, Xcelium ML could increase the coverage of P2 and P7 configurations, whereas ranking does not increase the original coverage.

Table V. Results produced by Ranking on mixed signal SoC

| Configuration | Original Runs | Original Runtime CPU hours | Ranked Runs | Ranked Regression Runtime CPU hours | Compression Factor in Runs | Coverage Regain | Compression Factor in Runtime |
|---|---|---|---|---|---|---|---|
| P1 | 297 | 0.200 | 87 | 0.057 | 3.41 | 100% | 3.51 |
| P2 | 415 | 0.562 | 115 | 0.189 | 3.60 | 100% | 2.97 |
| P3 | 2959 | 2.813 | 624 | 0.630 | 4.74 | 100% | 4.47 |
| P4 | 140 | 0.213 | 73 | 0.129 | 1.91 | 100% | 1.65 |
| P5 | 1193 | 1.336 | 217 | 0.232 | 5.49 | 100% | 5.76 |
| P6 | 10 | 0.007 | 10 | 0.007 | 1 | 100% | 1 |
| P7 | 110 | 0.074 | 78 | 0.062 | 1.41 | 100% | 1.19 |
| Total | 5124 | 5.205 | 1204 | 1.312 | 4.25 | 100% | 3.97 |

*C. Radar based SoC*

This SoC is an automotive electronics product for RADAR applications. It is an on-going project where the RTL design is written in SystemVerilog and the corresponding testbench is written in SystemVerilog-UVM. It has directed tests along with CRV based tests. As it is an on-going project, the ranking method and Xcelium ML are used on the testsuite at two stages of the development cycle. During the first stage, the testbench had 25 tests in total and 271 test runs in the original regression. As the development proceeded further, 10 tests were added to the testbench later. The final regression consisted of 301 test simulations.

The original regression had 271 runs but only 248 simulation runs passed and the remaining ones failed due to functional bugs. When the ML method was applied on this regression, it produced a condensed regression consisting of 117 runs. 15 simulation runs of the optimized regression failed indicating the functional errors. And also, this optimized regression was able to regain almost 100% of the original coverage. This can be seen in Table VI.

Table VI. Results produced by Xcelium ML on Radar based SoC

|          | **Original Runs**          | **Original Runtime CPU hours** | **Optimized Runs**         | **Optimized Runtime CPU hours** | **Compression Factor in Runs** | **Coverage Regain** | **Compression Factor in Runtime** |
|----------|----------------------------|-------------------------------|----------------------------|--------------------------------|--------------------------------|---------------------|-----------------------------------|
| Stage I  | 271<br>248 Passed<br>23 Failed | 0.162                         | 117<br>102 Passed<br>15 Failed | 0.070                          | 2.31                           | 99.77%              | 2.3                               |
| Stage II | 301<br>298 Passed<br>3 Failed  | 0.243                         | 112<br>89 Passed<br>23 Failed  | 0.079                          | 2.68                           | 108.42%             | 3.07                              |

As the testbench improves during the development, the tool was used on a new regression suite of 301 runs for optimization at the later stage. It had only 3 functional failures and the remaining 298 simulation runs passed. The optimized regression is 2.68 times shorter than the original regression and has 112 test runs. But interestingly, it regained more than 100% of the original coverage by hitting new coverbins and also executing the new blocks of the design code, thus improving the total coverage. This optimized regression has 23 failed runs, indicating the ability to discover a lot of new functional bugs during the simulation.

Ranking is performed on these regressions at the same stages. The results produced by ranking are tabulated in Table VII. It can be seen at both stages that, ranking produced similar compression factors as Xcelium ML.

Table VII. Results produced by Ranking on Radar based SoC

|          | **Original Runs**          | **Original Runtime CPU hours** | **Ranked Runs** | **Ranked Regression Runtime CPU hours** | **Compression Factor in Runs** | **Coverage Regain** | **Compression Factor in Runtime** |
|----------|----------------------------|-------------------------------|-----------------|------------------------------------------|--------------------------------|---------------------|-----------------------------------|
| Stage I  | 271<br>248 Passed<br>23 Failed | 0.162                         | 91              | 0.063                                    | 2.97                           | 100%                | 2.57                              |
| Stage II | 301<br>298 Passed<br>3 Failed  | 0.243                         | 101             | 0.074                                    | 2.98                           | 100%                | 3.28                              |

*D. Ranking vs Xcelium ML*

This section focuses on the comparison between ranking and Xcelium ML. Ranking produced 100% coverage regain for every project applied as it just selects the test and seed pairs from the input regression whereas Xcelium ML produced 99% or more coverage regain consistently with randomized regressions. These randomized regressions can exercise new random scenarios resulting in coverage regain of more than 100% at certain situations. Ranking can be very helpful, if the final goal is to produce direct simulation runs from the original regression. This technique is not recommended for exploration of the random space or exposing new bugs as it would not simulate any new scenarios. A graph is plotted for Coverage regains for ranking and Xcelium ML when applied on the various projects, which can be seen in Figure 3.

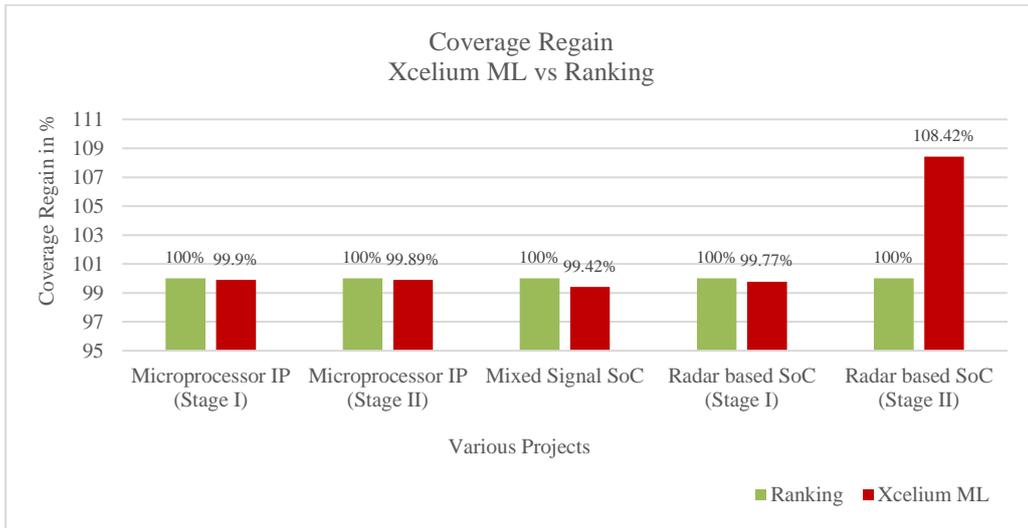

Figure 3. Comparison between Ranking & Xcelium ML with Coverage Regain

Also, optimization factor in runs was used to compare between ranking and Xcelium ML. The respective plot is shown in Figure 4. It can be seen that the compression factors are high for both of the techniques. Although ranking has good compression factors, it cannot improve total coverage or exercise new random scenarios. Xcelium ML produced around 3x shorter random regression consistently. The Microprocessor IP has higher compression factors for both ranking and Xcelium ML techniques because of a lot of redundant simulations inside the original regression which do not contribute for the final coverage. Both the techniques excel when applied on this particular project.

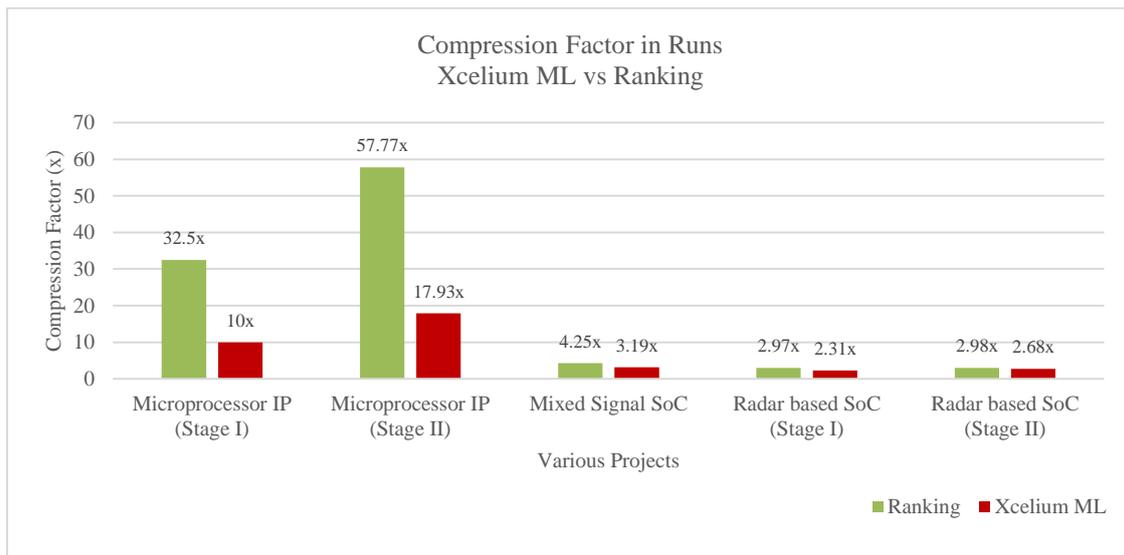

Figure 4. Comparison between Ranking & Xcelium ML with Compression Factor in Runs

Compression in CPU Runtime is used as a metric to compare the performances of both Xcelium ML and ranking techniques. A graph is plotted to see the compression factors for every project as shown in Figure 5. The factors for the microprocessor IP project are high as well, similar to the previous comparison in Figure 4. As mentioned earlier, most of the redundant simulations inside the original regressions are removed by both of the techniques. Here, during stage I of the microprocessor IP, Xcelium ML produced faster regressions than ranking. But during stage II, ranking produced faster regressions than Xcelium ML.

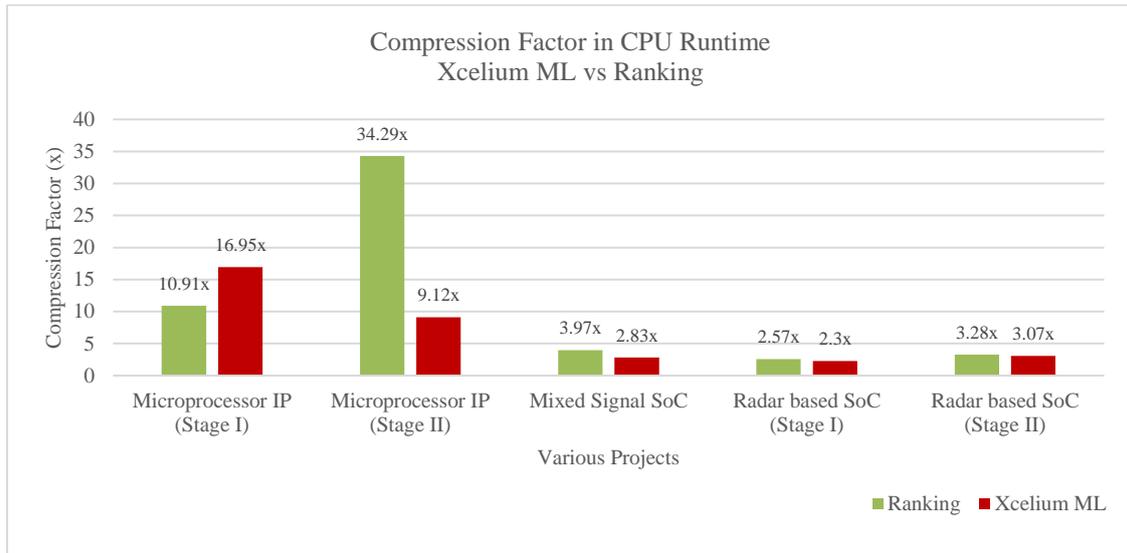

Figure 5. Comparison between Ranking & Xcelium ML with Compression Factor in CPU Runtime

## V. OUTLOOK - PROPOSED METHODOLOGY USING XCELIUM ML

Figure 6 depicts a proposed methodology for the verification process throughout a project timeline using the Xcelium ML tool. The simulation regression(s) is the input to the tool to produce a compressed and faster regression(s). This shorter regression can save simulation resources and time by using it regularly to find bugs in the design faster. After the optimized Xcelium ML regression is generated, two decisions are to be taken based on coverage regain and design/ testbench releases during the product development. If the optimized coverage regain is 99% or more, compressed Xcelium ML regression can be used regularly as a daily regression to exercise new scenarios. As it is a randomized regression, it can even expose some new bugs when it is run multiple times. In case the coverage regain is less than 99%, the goals for the Xcelium ML have to be set tighter and the optimization step has to be repeated until the coverage regain exceeds the 99% threshold. The Xcelium ML steps also have to be repeated if there are any significant changes inside the design or testbench (typically every week).

The Xcelium ML stage is shown at the right side of Figure 6. It consists of 3 main steps, namely, data collection, learning, generation. Data collection involves collecting the data of random control points, influential variable classes, and sensitive variables inside the testbench for each simulation. The learning step generates iterative learning models to determine which new constraints to include in a simulation to speed up the time to reach the original coverage. These learning models are used to produce the optimized and randomized regressions during generation step. The analysis step explains how control points influence coverage bins.

At present, the proposed methodology is being applied on the Radar SoC project regularly. The results shown in Table VI explains that the original regression in Stage I has 271 runs which are optimized to 117 runs using Xcelium ML. It was able to reproduce 99% or more coverage of the original regression, so this compressed ML regression was used regularly to detect bugs. There was a new release in the design and testbench of the project in Stage II, which has 301 runs in the regression. This new simulation regression is compressed to 112 runs, which exposed new design issues and increased the actual coverage.

This methodology enables the most efficient usage of Xcelium ML during the project development to save simulation resources and time.

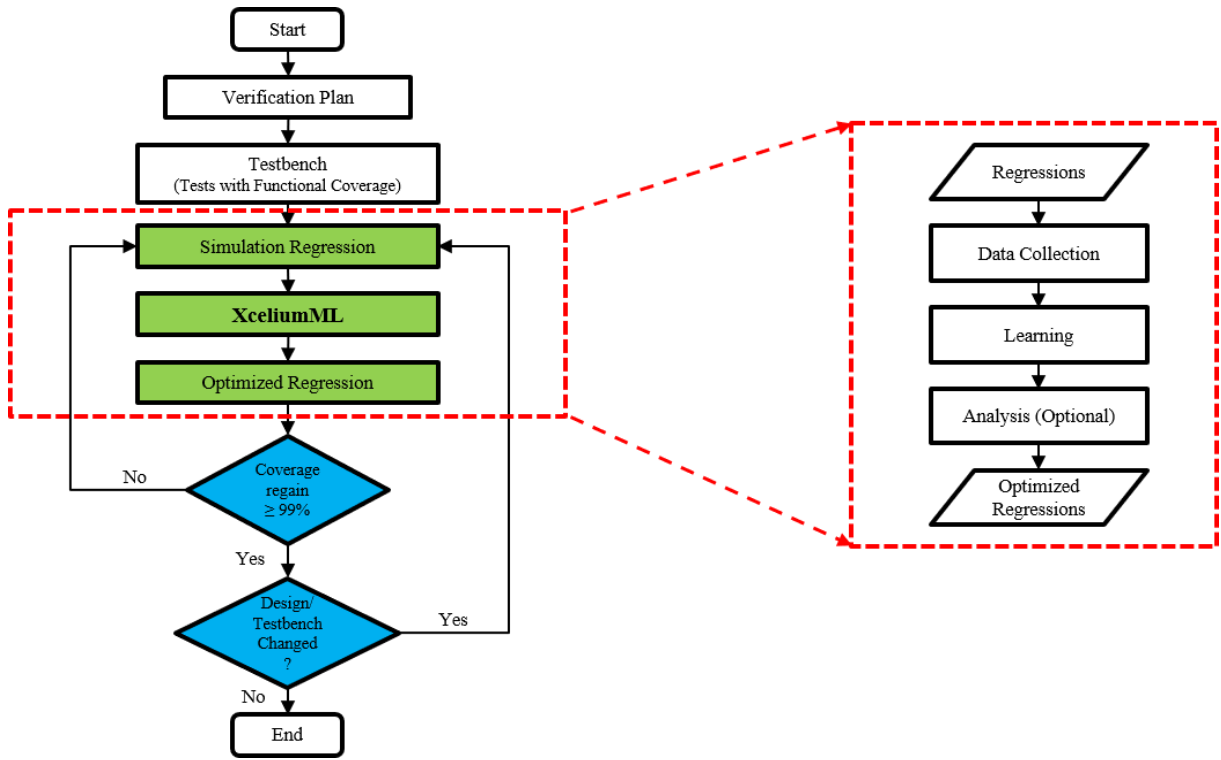

Figure 6. Proposed Methodology using Xcelium ML tool

## VI. CONCLUSION

Experimental results of Xcelium ML on various designs showed minimum of 3x compression factor, and 99% or more coverage regain consistently. At certain scenarios, for the optimized regressions, the compression factor was about 15 and the coverage regain is more than 100% which indicates the possibility to exercise the new scenarios in shorter time.

In general, the compression factors achieved by ranking are higher than those by Xcelium ML, whenever for most cases in a very similar range. At the same time, an optimized Xcelium ML regression remains a fully random with all advantages of random verification, first and foremost the ability to hit new scenarios and uncover new bugs.

The methodology can be used for production to optimize simulation regressions where the verification testbench uses randomized tests written in SystemVerilog. It can save simulation resources and CPU runtime by significant factors. It is currently being applied on multiple on-going projects to yield results on how the bugs can be exposed faster with the ML optimized regression in comparison with the original regression.